\documentclass[conference]{IEEEtran}
\IEEEoverridecommandlockouts
\usepackage{cite}
\usepackage{amsmath,amssymb,amsfonts, hyperref}
\usepackage{algorithmic}
\usepackage{graphicx}
\usepackage{textcomp}
\usepackage{xcolor}
\def\BibTeX{{\rm B\kern-.05em{\sc i\kern-.025em b}\kern-.08em
    T\kern-.1667em\lower.7ex\hbox{E}\kern-.125emX}}
    
\begin{document}

\title{Capturing Dynamics: The 4D Facial Expression Intensity Dataset}
% {\footnotesize \textsuperscript{*}Note: Sub-titles are not captured for https://ieeexplore.ieee.org  and

% \author{\IEEEauthorblockN{1\textsuperscript{st} Anonymous}
% \IEEEauthorblockA{
% \textit{Anonymous}\\
% \textit{Anonymous} \\
% Anonymous}
% }
\author{\IEEEauthorblockN{1\textsuperscript{st} Zesheng Wang}
\IEEEauthorblockA{
\textit{École Centrale Nantes, CNRS, LS2N, UMR 6004}\\
\textit{Nantes Université, France} \\
zesheng.wang@etu.univ-nantes.fr}
\and
\IEEEauthorblockN{2\textsuperscript{nd} Alexandre Bruckert}
\IEEEauthorblockA{
\textit{École Centrale Nantes, CNRS, LS2N, UMR 6004}\\
\textit{Nantes Université, France} \\
Alexandre.Bruckert@univ-nantes.fr}
\and
\IEEEauthorblockN{3\textsuperscript{rd} Pierre Lebreton}
\IEEEauthorblockA{
\textit{École Centrale Nantes, CNRS, LS2N, UMR 6004}\\
{\& CAPACITÉS SAS} \\
\textit{Nantes Université, France} \\
pierre.lebreton@univ-nantes.fr}
\and
\IEEEauthorblockN{4\textsuperscript{th} Patrick Le Callet}
\IEEEauthorblockA{
\textit{École Centrale Nantes, CNRS, LS2N, UMR 6004}\\
{\& Institut universitaire de France (IUF)}\\
\textit{Nantes Université, France} \\
Patrick.Le-Callet@univ-nantes.fr}
\and
\IEEEauthorblockN{5\textsuperscript{th} Yante Li}
\IEEEauthorblockA{\textit{Center for Machine Vision and Signal Analysis}\\
\textit{University of Oulu, Finland} \\
Yante.Li@oulu.fi}
\and
\IEEEauthorblockN{6\textsuperscript{th} Guoying Zhao}
\IEEEauthorblockA{\textit{Center for Machine Vision and Signal Analysis}\\
\textit{University of Oulu, Finland} \\
guoying.zhao@oulu.fi}
}
% \author{
% \IEEEauthorblockN{1\textsuperscript{st} Zesheng Wang}
% \IEEEauthorblockA{\textit{École Centrale Nantes, CNRS, LS2N, UMR 6004}\\
% \textit{Nantes Université, France} \\
% zesheng.wang@etu.univ-nantes.fr}%
% \and
% \IEEEauthorblockN{2\textsuperscript{nd} Alexandre Bruckert}
% \IEEEauthorblockA{\textit{École Centrale Nantes, CNRS, LS2N, UMR 6004}\\
% \textit{Nantes Université, France} \\
% Alexandre.Bruckert@univ-nantes.fr}%
% \and % --- 第一行结束 ---
% \IEEEauthorblockN{3\textsuperscript{rd} Pierre Lebreton}
% \IEEEauthorblockA{\textit{École Centrale Nantes, CNRS, LS2N, UMR 6004}\\
% {\& CAPACITÉS SAS} \\
% \textit{Nantes Université, France} \\
% pierre.lebreton@univ-nantes.fr}%
% \and
% \IEEEauthorblockN{4\textsuperscript{th} Patrick Le Callet}
% \IEEEauthorblockA{\textit{École Centrale Nantes, CNRS, LS2N, UMR 6004}\\
% {\& Institut universitaire de France (IUF)}\\
% \textit{Nantes Université, France} \\
% Patrick.Le-Callet@univ-nantes.fr}%
% \and % --- 第二行结束 ---
% \IEEEauthorblockN{5\textsuperscript{th} Yante Li}
% \IEEEauthorblockA{\textit{Center for Machine Vision and Signal Analysis}\\
% \textit{University of Oulu, Finland} \\
% Yante.Li@oulu.fi}%
% \and
% \IEEEauthorblockN{6\textsuperscript{th} Guoying Zhao}
% \IEEEauthorblockA{\textit{Center for Machine Vision and Signal Analysis}\\
% \textit{University of Oulu, Finland} \\
% guoying.zhao@oulu.fi}%
% }

\maketitle

\begin{abstract}
The estimation and analysis of facial expression intensity play a crucial role in affective communication and human-computer interaction. Previous research has primarily focused on detecting and estimating facial expression intensity from frame-level 2D representations. However, this limitation restricts a comprehensive understanding of real-world facial expressions, as they are inherently 3D and temporally continuous. This paper investigates the perception of facial expression intensity by introducing the 4D Facial Expression Intensity Dataset (4D-FEID). We employ a parametric face model and compile a total of 2,869 mesh sequences with controlled geometric variations, generating 4D data instances with diverse peak intensities and identity attributes. Using a Likert scale, we collect more than 90,000 subjective intensity perception ratings via a crowdsourcing platform. We explore various architectures and aggregation methods to establish baselines for episode intensity estimation on the new dataset, revealing that spatial-temporal graph models consistently outperform traditional frame-aggregation methods. In contrast to existing datasets that rely on 2D static imagery, the proposed 4D-FEID dataset provides the community with a unique and vital resource for investigating the perception of facial expression intensity through the use of dynamic 3D stimuli. By offering high-fidelity, spatio-temporally coherent facial data, 4D-FEID establishes a new foundation for research into more nuanced and naturalistic expression analysis, thereby addressing a gap in the current landscape of affective computing and human-computer interaction studies. The dataset is available at \href{https://zenodo.org/records/17152903?token=eyJhbGciOiJIUzUxMiJ9.eyJpZCI6IjNhYjI5ZWRkLTk5NDYtNDUwMi04NzQ1LTBlOWE2ODdlYmU1YiIsImRhdGEiOnt9LCJyYW5kb20iOiIwYTBmZDc4MGMxZWI1MmU2ZjEyMmZlZjFhM2ViNzM3NyJ9.PBN44DuHQSHlFL12MXiUIkMxAI78u-b1v6Zs_b9JeribfTzyhxTz58lTrzsF6W70sVv33gHCWADs4qjUwrbS8Q}{link}.
\end{abstract}


\section{Introduction}
Facial expression constitutes a vital non-verbal channel for conveying internal emotional states, attracting significant research attention. Recent years have witnessed growing interest in fine-grained facial expression analysis, particularly expression intensity, which plays a critical role in discriminating subtle emotional variations \cite{zhao2016facial, valstar2015fera}.
Several datasets incorporating facial expression intensity have been introduced \cite{valstar2015fera, mavadati2013disfa, zhang2013high}. These efforts typically focus on intensity perception across temporal phases of a single expression, emphasizing frame-level facial action unit (AU) recognition and intensity estimation from static images. However, the perception of AU intensity from 2D images is often hindered by illumination artifacts and viewpoint constraints, which may introduce bias and impair robustness. 
% Furthermore, annotating individual AUs and intensity requires certified Facial Action Coding System (FACS) experts \cite{ekman2002facial}, inherently limiting dataset scalability.
Furthermore, the need for certified FACS experts \cite{ekman2002facial} to annotate individual AUs severely limits dataset scalability.

In contrast, humans rely on dynamic 3D visual cues to perceive expression intensity in daily interactions. 
% The transition to 4D (dynamic 3D) data is crucial, as it captures the nuanced spatio-temporal deformation of facial geometry that 2D projections fail to represent. Such 4D representations provide the high-fidelity structural information necessary for modeling the subtle progression of emotional states.
Transitioning to 4D data is crucial; unlike 2D projections, it captures the high-fidelity spatio-temporal deformations necessary for modeling subtle emotional progressions.
However, existing 4D facial datasets \cite{COMAECCV18} primarily focus on categorical labels or identity reconstruction, often overlooking the continuous spectrum of expression intensity. To bridge this gap, we propose the construction of a 4D facial mesh dataset that facilitates intensity estimation under more natural and ecologically valid conditions.

To address the lack of intensity variability in existing mesh corpora, we leverage the 3D morphable model, FLAME \cite{li2017learning}, to generate a large-scale, 4D dataset with rigorously controlled expression intensities. 
% By systematically sampling the expression manifold at multiple granular levels, we produce personalized 4D sequences that decouple subtle intensity shifts from identity shape variations. 
By sampling the expression manifold derived from high-fidelity 4D captures/scans, we generate personalized 4D sequences that decouple intensity shifts from identity variations, ensuring both physical realism and identity diversity.
This approach overcomes the scarcity of intensity labels in current 4D benchmarks. Furthermore, unlike traditional frame-level annotations that isolate expression phases, human perception of dynamic expressions emerges from the global spatio-temporal evolution of 3D geometry \cite{muukkonen2020spatio}. Accordingly, we define facial expression intensity as an episode attribute, reflecting the holistic perceived score assigned by observers to the entire sequence. This paradigm shift not only aligns with human affective cognitive processes but also bypasses the need for complex, AU expert coding.

We make three key contributions:
\begin{itemize}
\item First, we introduce 4D-FEID, a large-scale 4D facial expression intensity dataset comprising 2,869 mesh sequences. Each sequence is paired with its underlying FLAME parameters \cite{li2017learning}, offering a rich resource for both geometric and parametric analysis.
\item Second, we propose a novel continuous intensity labeling scheme based on a 5-point Likert scale (1–5), and collect over 90,000 human annotations, ensuring the statistical reliability of the perceived intensity.
\item Third, we conduct extensive evaluations of three distinct types of methods on the proposed dataset, establishing the first benchmark for 4D facial expression intensity estimation and providing a foundation for future research in dynamic 3D facial behavior analysis.
\end{itemize}

\section{RELATED WORK}
\label{sec:relatedwork}
\subsection{Perceived Intensity and Estimation}
\label{ssec:perceived_estimation}
The perception of facial expression intensity traditionally relies on rating scales, a cornerstone psychometric tool. Ekman et al. \cite{ekman1987universals} pioneered such scales to examine cross-cultural intensity judgments, identifying both universal trends and cultural variations. Standardized ordinal scales—such as the five-level (A–E) metric in the FACS \cite{ekman2002facial} and the annotations in DISFA \cite{mavadati2013disfa}—have been widely adopted for action unit (AU) intensity encoding. Nevertheless, intensity assessment from static images remains vulnerable to illumination artifacts and sensor noise \cite{ekman2002facial}.

To address these challenges, computational methods for AU intensity estimation have evolved significantly. Early works framed the task as multi-class classification \cite{delannoy2008automatic} or ordinal regression \cite{chang2013intensity,zhao2016facial}. Recent advances leverage deep learning to improve robustness and accuracy: Gudi et al. \cite{gudi2015deep} proposed a CNN for joint AU detection and intensity prediction, while Walecki et al. \cite{walecki2017deep} introduced structured learning to model AU relationships and alleviate data imbalance. Further innovations include heatmap-based regression for spatially-aware intensity estimation \cite{fan2020facial} and self-supervised pre-training strategies that enhance generalization in data-scarce scenarios \cite{ma2024facial}.

\subsection{Relevant Datasets}
\label{ssec:relevant}

\begin{table}[bp]
\caption{The previous expression-intensity and 4D-expression datasets}
\begin{center}
\begin{tabular}{lcll}
\hline
% Dataset    & Modality & Perceptual & Annotations \\ \hline
% MultiEmoVA\cite{mou2015group} & 2D       & static              & Arousal               \\
% HAPPEI \cite{dhall2015automatic}    & 2D       & static              & Happiness Intensity   \\
% DISFA\cite{mavadati2013disfa}      & 2D       & static              & 12 AUs Intensity               \\
% BP4D\cite{zhang2014bp4d}       & 4D       & static              & 2 AUs Intensity          \\
% 4DME\cite{li20224dme}       & 4D       & dynamic              & None Intensity           \\
% 4D-FEID    & 4D       & dynamic             & Expression Intensity  \\ \hline
Dataset & Modality & Granularity & Intensity Type \\ \hline
MultiEmoVA \cite{mou2015group} & 2D & Frame-level & Arousal \\
HAPPEI \cite{dhall2015automatic} & 2D & Frame-level & Happiness \\
DISFA \cite{mavadati2013disfa} & 2D & Frame-level & AUs-based \\
BP4D \cite{zhang2014bp4d} & 4D & Frame-level & AUs-based \\
4DME \cite{li20224dme} & 4D & Sequence-level & None (Categorical) \\ 
\textbf{4D-FEID (Ours)} & \textbf{4D} & \textbf{Sequence-level} & \textbf{Global Expression} \\ \hline
\end{tabular}

\label{tab:1}
\end{center}
\end{table}

Despite these advancements, most existing algorithms are strictly tailored for 2D representations. The recent shift towards 3D and 4D facial analysis has prompted the adoption of geometric deep learning, such as point-based networks (e.g., PointNet++ \cite{qi2017pointnet++}), graph convolutional networks (GCNs) \cite{COMAECCV18}, and 3D spatio-temporal architectures \cite{yan2018spatial}, to capture complex geometric deformations. However, the application of these advanced spatio-temporal models to continuous expression intensity estimation remains underexplored, primarily due to the lack of dedicated 4D datasets with comprehensive intensity annotations.
%Two widely used datasets for facial expression intensity estimation are DISFA \cite{mavadati2013disfa} and BP4D \cite{zhang2014bp4d}. DISFA provides 27 videos with frame-level AU intensity labels annotated by FACS experts. While BP4D captures 3D dynamic geometry, its intensity annotations are predominantly derived from 2D video sequences and limited to sparse AUs. As summarized in Table \ref{tab:1}, existing datasets predominantly rely on 2D imagery and lack continuous 4D expression dynamics. In contrast, the proposed 4D-FEID dataset introduces \textbf{high-fidelity 4D facial mesh sequences}, \textbf{large-scale continuous intensity ratings}, and \textbf{systematically controlled expression variations}, thereby addressing critical gaps in spatial and temporal modeling of expressive behavior.

Considerable efforts have been devoted to capturing authentic dynamic facial expression meshes \cite{cheng20174dfab, principi2023florence}. However, due to the complexity of acquisition protocols, there remains a lack of research regarding expression intensity.
Two benchmark datasets frequently employed for intensity estimation are DISFA \cite{mavadati2013disfa} and BP4D \cite{zhang2014bp4d}. DISFA provides extensive frame-level AU intensity labels for 2D video sequences; however, its reliance on 2D texture makes it susceptible to lighting and pose variations. While BP4D incorporates 3D dynamic geometry, its intensity annotations are predominantly derived from 2D perspectives and remain confined to a sparse set of Action Units, and are limited to frame-level granularity, which ignores the holistic temporal evolution of expressions. As synthesized in Table \ref{tab:1}, existing corpora are largely constrained by 2D projections and the absence of holistic 4D expression dynamics. In contrast, the proposed 4D-FEID dataset bridges these gaps by providing \textbf{high-fidelity 4D facial mesh sequences}, \textbf{sequence-level continuous intensity ratings}, and \textbf{systematically controlled expression variations}. These features collectively enable a more robust framework for the spatio-temporal modeling of expressive human behavior.

\section{THE 4D-FEID DATASET}
\label{sec:method}
The 4D-FEID dataset consists of three aligned components: (i) 4D mesh sequences, (ii) corresponding FLAME parameters, and (iii) rendered frontal-view videos. The raw mesh data are sourced from D3DFACS \cite{cosker2011facs}, which captures facial-expression mesh sequences at 60 fps. We represent a mesh sequence as 
$\mathcal{X} =[X_1, \cdots , X_n],$
where n denotes the number of sequences, and each sequence $X_i$ is represented by follow:
% \begin{equation}
% X_i = [F_i^T, V_i^T], \label{1}
% \end{equation}
\begin{equation}
X_i = (F, V_i), \label{1}
\end{equation}
% where $F$ denotes the fixed set of triangular faces, $V_i^T$ contains the 3D coordinates of all vertices at the $T$ frames, and 
% $i\in \left \{ 1,\cdots ,n \right \} $. 
where $F$ denotes the fixed topology of triangular faces across all frames, and $V_i \in \mathbb{R}^{T \times N_v \times 3}$ contains the 3D coordinates of all $N_v$ vertices across $T$ frames for the $i$-th sequence.
The vertex coordinates at a specific frame $t$ are represented as $V_i^t$.
% The vertex coordinates for each sequence are represented as
% $V_i^T = [V^T_{ix} , V^T_{iy} ,V^T_{iz}]\in R^{T\times 3}$, 
% containing the positional information across T frames. 

% Following FLAME \cite{li2017learning}, we register each mesh frame to the FLAME model through single-frame fitting, obtaining:
% \begin{equation}
% \vec{\beta_i^t},\vec{\theta_i^t},\vec{\psi_i^t} =\mathcal{M}_{w_{i}^t}(F_{i}^t,V_{i}^t),t\in \left \{ 1,\cdots T \right \} , \label{2}
% \end{equation}
% where $\mathcal{M}$ denotes the registration model with parameters $w$, and $\vec{\beta_i^t},\vec{\theta_i^t},\vec{\psi_i^t}$ represent the identity shape, head pose, and expression parameters for subject $i$ at frame $t$, respectively.
Following FLAME \cite{li2017learning}, we register each mesh frame to the FLAME model. 
% 审稿人补充：解决了单帧拟合带来的Identity Jittering问题
To prevent identity jittering across temporal sequences, we first optimize the identity shape parameter $\vec{\beta_i}$ on the neutral onset frame and keep it constant. Subsequently, we estimate the frame-wise head pose and expression parameters via single-frame fitting:
\begin{equation}
\vec{\theta_i^t},\vec{\psi_i^t} = \mathcal{M}_{w_{i}^t}(F, V_{i}^t | \vec{\beta_i}), \quad t \in \{1,\cdots T\}, \label{2}
\end{equation}
where $\mathcal{M}$ denotes the registration model with parameters $w$, and $\vec{\beta_i},\vec{\theta_i^t},\vec{\psi_i^t}$ represent the identity shape, head pose, and expression parameters for subject $i$ at frame $t$, respectively.

\begin{figure}[!tb]
  \centering
  \centerline{\includegraphics[width=8.5cm]{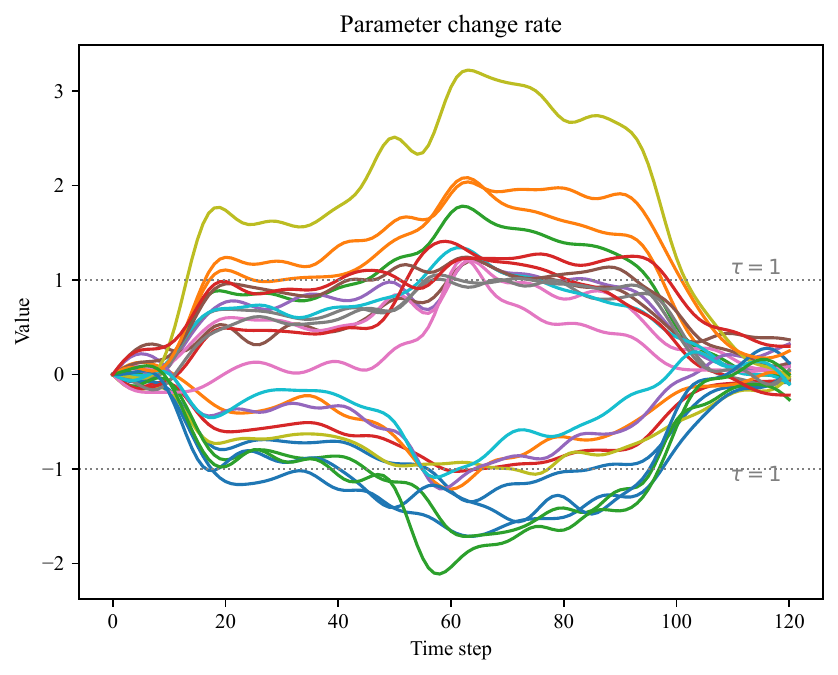}}
\caption{Example of intensity-based pruning on a sample sequence from
the dataset. Each colored curve corresponds to one of the preserved expression-parameter dimensions.}
\label{fig:1}
\end{figure}

Facial expressions result from the subtle and coordinated activation of multiple facial muscles and may appear as brief micro-expressions lasting less than 500 ms \cite{li2022cas}. 
By modulating the facial expression parameters $\vec{\psi}$ after registration, the model enables controllable editing of 4D facial expressions.
To preserve temporal continuity and prevent interpolation artifacts during registration from distorting these dynamics, we apply a temporal low-pass filter to the recovered expression parameters $\vec{\psi}$. 
The filtering operation is defined as:
\begin{equation}
\hat \psi_{i} =lowpass(\vec{\psi_i}), \label{3}
\end{equation}
where $lowpass$ denotes a 5th-order Butterworth low-pass filter applied using forward-backward filtering to eliminate phase lag and ensure zero-phase distortion. 
The filtering result is shown in Figure \ref{fig:1}, where the post-processed parameters demonstrate a significant reduction in discontinuities induced by spike noise, while preserving the underlying expression dynamics.
% The filter significantly reduces expression discontinuities caused by parameter spike noise while preserving the underlying expression dynamics.

% To enhance the variability of expression intensity, we amplify the registered expression parameters $\hat \psi$ by scaling their magnitudes. However, spontaneous facial motion inherently includes subtle, physiologically plausible tremors (e.g., natural lip quivers).
% To preserve such micro-movements, we compute the frame-wise deviation from the onset value for each dimension of $\hat \psi$, and retain all components whose absolute deviation remains above an adaptive threshold:

% \begin{equation}
% \tau =\left \| \hat{\psi }^{t_{max}}-\hat{\psi }^{t_{onset}}\right \|_2, \tau \in R^{100\times T}, \label{3}
% \end{equation}
To enhance the variability of expression intensity, we amplify the registered expression parameters $\hat \psi$ by scaling their magnitudes. However, spontaneous facial motion inherently includes subtle, physiologically plausible tremors (e.g., natural lip quivers). To preserve such micro-movements, we compute the channel-wise absolute deviation from the onset value for each expression dimension $c$, and retain all components whose absolute deviation remains above an adaptive threshold:
% 审稿人修改：修正了数学表达，去掉了L2范数，改为通道级绝对差值，修复了标号
\begin{equation}
\tau_c = |\hat{\psi}_c^{t_{max}} - \hat{\psi}_c^{t_{onset}}|, \quad \vec{\tau} \in \mathbb{R}^{d}, \label{4}
\end{equation}
where $d$ is the dimension of the FLAME expression parameter. In this work, we select the parameter subset satisfying $\tau_c \ge 1$ by analyzing the rate of change across all expression parameters. As illustrated in Figure \ref{fig:1}, only channels with deviations exceeding this threshold are scaled, thereby ensuring that intrinsic physiological tremors remain unaltered. Then, we scale the selected expression parameters across all frames $t$ according to the following formula:
\begin{equation}
\widetilde{\psi}_c^t = 
\begin{cases}
\hat\psi_c^t + \bigl(\hat\psi_c^t - \hat\psi_c^{t_{onset}}\bigr) \times (scale - 1), & \text{if } \tau_c \ge 1\\[4pt]
\hat\psi_c^t, & \text{if } \tau_c < 1 , \label{5}
\end{cases}
\end{equation}

% In this work, we select the median parameter set $\tau \ge 1$ by analyzing the rate of change across all expression parameters. As illustrated in Figure \ref{fig:1}, only deviations exceeding $\tau$ are scaled, thereby ensuring that intrinsic physiological tremors remain unaltered. Then, we scale the expression parameters obtained for each sequence using the above method according to the following formula:
% \begin{equation}
% \widetilde{\psi}= 
% \begin{cases}
% \hat\psi^t(\tau)+\bigl(\hat\psi^t(\tau)-\hat\psi^1(\tau)\bigr)\times scale, & \tau\ge 1\\[4pt]
% \hat\psi(\tau), & \tau<1 , \label{4}
% \end{cases}
% \end{equation}
% where $\psi(\tau)$ denotes the channel-dimensional parameters of $\psi$ under condition $\tau$. 
% In this work, we adopt a step size of 0.5 for the scale factors and additionally include two attenuation factors that reduce expression intensity. The scaling factors are set as follow:
% \begin{equation}
% \begin{matrix}  scale\in
% \bigl\{0.1,\,0.3,\,0.5,\,1.5,\,2.0,\\
% \quad\quad\quad\quad\quad\,2.5,3.0,\,3.5,\,4.0,\,4.5,\,5.0\bigr\}.
% \end{matrix}
% \end{equation}
In this work, we adopt a step size of 0.5 for the scale factors and additionally include two attenuation factors that reduce expression intensity. The scaling factors are empirically set as $scale \in \{0.1, 0.3, 0.5, 1.5, 2.0, 2.5, 3.0, 3.5, 4.0, 4.5, 5.0\}$.

\begin{figure}[!tb]
  \centerline{\includegraphics[width=8.5cm]{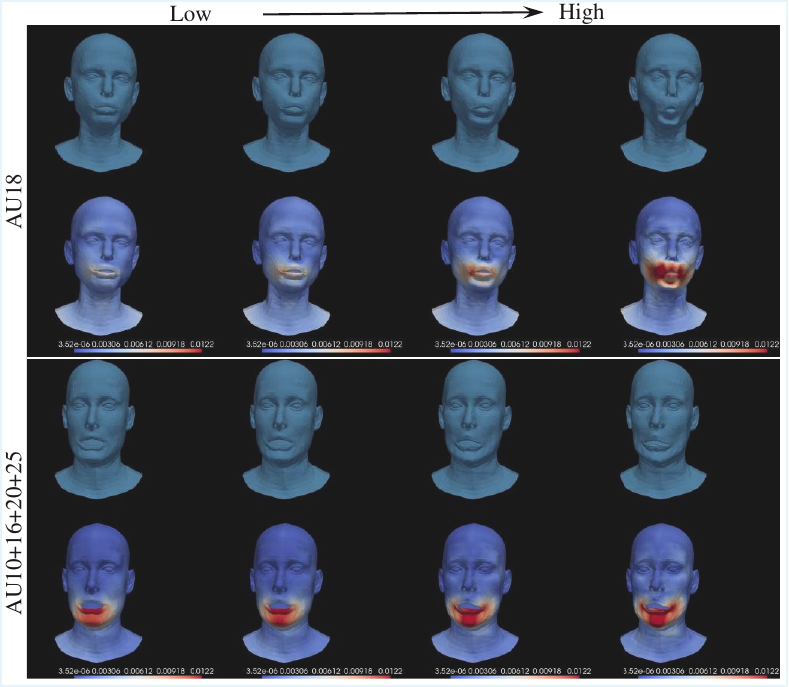}}
\caption{Examples of rendering outputs, displaying peak-frame results of various facial expressions under progressively increasing intensities. The overlaid heatmaps qualitatively verify that the expression intensity strengthens from left to right.}
\label{fig:2}
\end{figure}

As no prior work has systematically evaluated the naturalness of dynamically generated 3D facial expressions, we manually filter the synthesized sequences to eliminate unnatural artifacts.

Specifically, sequences exhibiting severe mesh self-intersections or physiologically implausible vertex distortions around the lip and eye regions were discarded. This rigorous quality-control process results in a final 4D facial expression dataset comprising 2,869 distinct sequences that exhibit a wide range of expression intensities.
% As no prior work has systematically evaluated the naturalness of dynamically generated 3D facial expressions, we manually filter the synthesized sequences to eliminate unnatural artifacts caused by abrupt intensity variations. This process results in a final 4D facial expression dataset comprising 2,869 distinct sequences that exhibit a wide range of expression intensities.

We compute the frontal-view camera pose from the head pose parameters $\vec{\theta}$, position a headlight at the same location, and introduce two additional fixed scene lights. Each facial sequence is subsequently rendered into a frontal-view video with a resolution of $1280\times960$ at 60 FPS. As shown in Figure \ref{fig:2}, we provide exemplars of $AU8$ and the combined AU$10+16+20+25$ from our dataset. For each expression, we render the facial geometry at the peak frame across sequences with varying intensities. To emphasize intensity differences, we compute the per-vertex Euclidean displacement between the 4D mesh at the peak expression frame and the onset frame. This displacement is then visualized as a color-coded heatmap texture applied to the mesh surface.

\subsection{Subjective Study}
\label{ssec:subjective}
% We recruited participants from Prolific \cite{prolificProlificEasily}, and subjective ratings were collected via a crowdsourcing service. 
% Firstly, we define a 5-point Likert scale with anchors: [Trace, Slight, Marked \& Pronounced, Severe \& Extreme, Maximum]. Before the experiment, the participants completed a brief tutorial to familiarize themselves with the procedure. Each stimulus consisted of a rendered video. After viewing a stimulus, participants rated the perceived expression intensity using the interface illustrated in Figure \ref{fig:3}. 

Firstly, we defined a 5-point Likert scale with anchors: [Trace, Slight, Marked \& Pronounced, Severe \& Extreme, Maximum]. These anchors are intentionally aligned with the five-level (A–E) intensity metric utilized in the established FACS \cite{ekman2002facial}, bridging global intensity perception with traditional expert coding. 

\begin{figure}[htb]
  \centerline{\includegraphics[width=8.5cm]{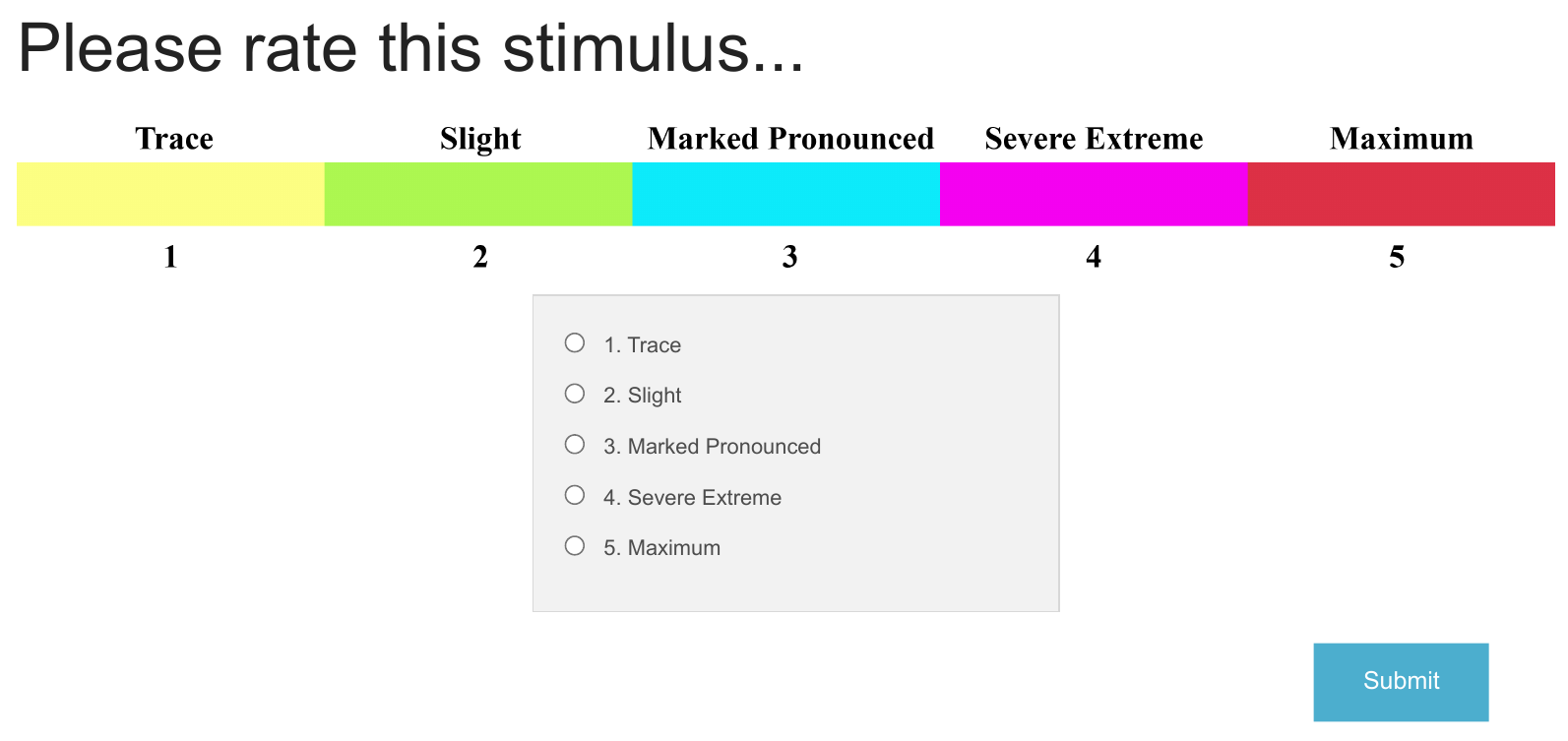}}
\caption{Interface of subjective experiment.}
\label{fig:3}
\end{figure}

Before the experiment, the participants completed a brief tutorial to familiarize themselves with the procedure. Each stimulus consisted of a rendered video. After viewing a stimulus, participants rated the perceived expression intensity using the interface illustrated in Figure \ref{fig:3}.

The experiment was organized into 14 sessions, each containing approximately 200 videos. Videos generated from the same 4D sequence were grouped within the same session. Each video was presented once, and each session lasted around 15 minutes. Every sequence was evaluated by at least 32 unique raters, resulting in a total of 91,584 ratings.

\subsection{Subjective Data Analysis}
\label{ssec:analysis}
The predicted intensities are mapped to a 5-point scale on 1 (minimum intensity) to 5 (maximum intensity). To quantify inter-observer reliability, we computed the intra-class correlation coefficient (ICC) separately for each experimental session. The resulting ICC values, ranging from 0.96 to 0.98 (see Table \ref{tab:2}), significantly exceed the conventional threshold of 0.80, indicating excellent agreement among raters \cite{cohen2013statistical}. In accordance with the ITU-T P.910 \cite{ITUP910} standard, we subsequently averaged the ratings across all observers to derive the ground-truth mean opinion score (MOS) for each sequence.

\begin{table}[htbp]
\caption{Inter-observer consistency measured by ICC for each session.}
\begin{center}
\begin{tabular}{|c|c|c|c|}
\hline
Session & Reliability(ICC) & Session & Reliability(ICC) \\ \hline
1       & 0.97             & 8       & 0.97             \\
2       & 0.97             & 9       & 0.98             \\
3       & 0.98             & 10      & 0.97             \\
4       & 0.98             & 11      & 0.97             \\
5       & 0.97             & 12      & 0.97             \\
6       & 0.97             & 13      & 0.97             \\
7       & 0.98             & 14      & 0.96             \\ \hline
\end{tabular}
\label{tab:2}
\end{center}
\end{table}

\section{EXPERIMENTS}
\label{sec:experiments}

\subsection{Experimental Setup}
\label{ssec:setup}
% 增加同类实验和泛化性实验，泛化性实验是数据集中抽取出来与训练集完全不同的3D模型shape和AU是训练阶段完全没见过shape和AU组合。
% We benchmark the task of expression intensity estimation on the proposed dataset. Leveraging both the rendered videos and the underlying 4D meshes, we evaluate three families of baseline methods:
We benchmark the task of expression intensity estimation on the proposed dataset. To rigorously evaluate the models, we partition the dataset into three subsets: a training set, a test set (following the same distribution), and a cross-identity generalization set. The generalization set specifically contains 4D shapes and AU combinations that are entirely excluded from the training phase, ensuring a robust assessment of the models' ability to handle unseen identities and expressions. Leveraging both the rendered videos and the underlying 4D meshes, we evaluate three families of baseline methods:

\begin{itemize}
\item  \textbf{4D-Vertex}: 
%This geometry-based baseline uses the raw XYZ coordinates of all mesh vertices at each time step. These are concatenated into a high-dimensional feature vector, from which the intensity is regressed after spatio-temporal modeling.
A geometry-based model that projects raw $XYZ$ vertex coordinates into a latent space via $1\times1$ convolutions. Global pooling ensures permutation invariance, followed by a 2-layer Bi-LSTM for temporal modeling.

\item \textbf{Video-CNN}: 
%A frame-level ResNet backbone extracts spatial features from sampled frames. A Bidirectional LSTM (Bi-LSTM) is then employed to capture the spatio-temporal dynamics, followed by a multi-layer perceptron (MLP) to regress the intensity score.
A ResNet-18 backbone extracts frame-level features \cite{gudi2015deep}. A Bi-LSTM with a temporal attention mechanism aggregates hidden states, prioritizing the expression apex via learned importance weights.

\item \textbf{FLAME-based}: 
%To evaluate the effectiveness of parametric models, we use only the FLAME expression coefficients as input for end-to-end intensity regression.
A parametric baseline using 100-D FLAME expression coefficients. It employs 1D convolutions (kernel size=3) to capture motion gradients, followed by an LSTM and MLP regressor.

\end{itemize}

\subsection{Implementation Details}
\label{ssec:implementation}

% \noindent \textbf{4D-Vertex}: For the raw 4D geometry, we employ a $1\times1$ convolutional layer which projects the 3D coordinates ($X, Y, Z$) vertices into a higher-dimensional latent space. We then apply global pooling across the vertex dimension to obtain a permutation-invariant spatial representation for each time step, which is subsequently processed by a $2$-layer Bi-LSTM to model the facial dynamics.

% \noindent \textbf{Video-CNN}: We utilize a ResNet-18 backbone to extract frame-level feature following \cite{gudi2015deep}. To address the varying significance of frames in an expression sequence, we implement a temporal attention layer consisting of two fully-connected layers and a Tanh activation. This layer computes a normalized weight for each frame, allowing the model to aggregate the Bi-LSTM hidden states into a single context vector that emphasizes the expression apex.

% \noindent \textbf{FLAME-based}: This model processes the $100$-dimensional FLAME expression coefficients. We employ a multi-layer 1D convolutional block with a kernel size of $3$ to capture local temporal correlations and motion gradients within the coefficient sequence. The output is then fed into a standard LSTM and a three-layer MLP regressor.

%\noindent \textbf{Data Partitioning and Training}:
We adopt a rigorous data splitting strategy to evaluate both model performance and robustness. The dataset is first randomly partitioned into a training set ($80\%$) and a test set ($20\%$). To further assess the generalization capability, we curate a dedicated cross-identity generalization set. This set consists of specific 3D shape identities and AU-intensity combinations that are entirely excluded from the training phase, ensuring the model is tested on unseen morphological structures and expression patterns.

To ensure a fair comparison, all benchmark models are trained under identical conditions for $100$ epochs using the Adam optimizer. To standardize the varying lengths of expression sequences, we apply a uniform temporal sampling strategy, fixing the data sampling rate $T$ at $32$ frames per sequence. We employ the Mean Squared Error ($L_2$) as the loss function with a constant learning rate of $1 \times 10^{-3}$.
% To ensure a fair comparison, all benchmark models are trained under identical conditions for $100$ epochs using the Adam optimizer. The data sampling rate $T$ is fixed at $32$. We employ an $L_2$ loss function with a fixed learning rate of $1 \times 10^{-3}$.

\subsection{Metrics}
\label{ssec:metrics}
%To evaluate the objective estimation performance, we report the mean absolute error (MAE), intra-class correlation coefficient (ICC), and Pearson correlation coefficient (PCC) on the test set.
To provide a comprehensive evaluation of the proposed methods, we assess their performance from multiple perspectives, including absolute estimation accuracy, linear correlation, and temporal consistency. Specifically, the following metrics are reported on the test set:

\begin{itemize} 
\item \textbf{Mean Absolute Error (MAE)}:
% Used to quantify the absolute precision of the intensity estimation by measuring the average magnitude of residuals between the predicted values and ground truth.
To quantify absolute prediction precision.

\item \textbf{Pearson Correlation Coefficient (PCC)}:
%Employed to evaluate the linear relationship and trend alignment, reflecting how well the model captures the dynamic fluctuations of expression intensity over time.
To evaluate linear trend alignment.

\item \textbf{Intra-class Correlation Coefficient (ICC)}: 
%The ICC(3,1) variant, which assesses the consistency and reliability of the model's predictions relative to the continuous labels, is a standard practice in affective computing benchmarks.
Specifically, the ICC(3,1) variant was used to measure the consistency and reliability of predictions relative to human-annotated labels.
\item \textbf{Concordance Correlation Coefficient (CCC)}:
%Adopted as a unified metric that simultaneously accounts for both precision (correlation) and accuracy (closeness to the 45-degree line), providing a holistic measure of the model's estimation quality.
To provide a holistic measure that combines both precision and accuracy.

\end{itemize}
\subsection{Results and Discussion}
\label{ssec:results}
% 讨论模型的失效和成功。说明微妙的表情变化很难像人类观察者一样感知。从而针对video-cnn和4d-vertex方法采用差异度感知学习。添加两个表，第一个表说明模型学习结果和对比人眼的ICC结果讨论；第二个表说明模型优缺点和差异学习模型的局限性。
% The strong performance of the FLAME-based baseline (Table 3), which achieves an MAE of 0.07, ICC of 0.80, and PCC of 0.94 on the test set, underscores both the feasibility and importance of accurate 4D expression-intensity estimation—a task critical to affective computing and human behavior understanding. In contrast, neither the Video-CNN nor the 4D-Vertex model reduces the training loss below an empirical failure threshold ($\mathcal{L}_{train} > 0.7$ after 50 epochs), and their quantitative results are omitted. The pronounced performance gap highlights the inherent challenge of this task: unlike static-image recognition, dynamic 3D expression-intensity estimation requires joint modeling of intricate spatial deformations and nuanced temporal dynamics, posing a substantially more complex learning problem. The moderate agreement attained by the FLAME-based model confirms the value of the proposed benchmark, while the failure of alternative architectures emphasizes that effective representation of 4D expressive behavior remains a non-trivial and open research challenge.
We observed that both the Video-CNN and 4D-Vertex models encountered significant optimization challenges when using raw data as input. Neither model could effectively reduce the training loss below an empirical failure threshold ($\mathcal{L}_{train} > 0.7$ after $50$ epochs). 
% This failure highlights a core challenge in 4D affective computing: subtle expression variations are often overshadowed by dominant identity-related spatial information, making it difficult for standard architectures to perceive nuanced intensity shifts as human observers do.
This failure highlights a core challenge in affective computing: subtle expression variations are easily overshadowed by the dominant static geometry of individual identities. Standard architectures struggle to isolate these nuanced intensity shifts from raw absolute coordinates as human observers do.

To address this, we implement a difference-aware learning strategy for the Video-CNN and 4D-Vertex methods. Specifically, instead of using raw frames or coordinates, the models are trained on the residuals between the current expression and the initial neutral state of the same identity. This approach effectively filters out static morphological noise and forces the networks to focus on dynamic deformations. The quantitative results of this strategy are reported in Table \ref{tab:3} for the standard test set and Table \ref{tab:4} for the cross-identity generalization set.

\begin{table}[htbp]
\caption{Test-set performance of the baseline methods in the standard set. Symbols $\uparrow$ and $\downarrow$ indicate that higher and lower values are better, respectively.}
\begin{center}
\begin{tabular}{lcccc}
\hline
Model       & MAE$\downarrow$ & ICC$\uparrow$ & PCC$\uparrow$ & CCC$\uparrow$\\ \hline
FLAME-based & 0.192            & 0.946         & 0.946          & 0.943\\ 
Video-CNN   & 0.289            & 0.960         & 0.960          & 0.911\\  %频流感受到4D的变化...
4D-Vertex   & \textbf{0.139}            & \textbf{0.975}          & \textbf{0.975}          & \textbf{0.974}\\ \hline  %说明人类观察者可以通过观察视
\end{tabular}
\label{tab:3}
\end{center}
\end{table}

\begin{table}[htbp]
\caption{Test-set performance of the baseline methods in the cross-identity set. Symbols $\uparrow$ and $\downarrow$ indicate that higher and lower values are better, respectively.}
\begin{center}
\begin{tabular}{lcccc}
\hline
Model       & MAE$\downarrow$ & ICC$\uparrow$ & PCC$\uparrow$ & CCC$\uparrow$\\ \hline
FLAME-based & 0.204            & 0.942          & 0.942         & 0.939\\ 
Video-CNN   & 0.282            & 0.962          & 0.961         & 0.917\\  
4D-Vertex   & \textbf{0.146}            & \textbf{0.973}          & \textbf{0.973}         & \textbf{0.970}\\ \hline %说明人类观察者可以通过观察视频流感受到4D的变
\end{tabular}
\label{tab:4}
\end{center}
\end{table}

The quantitative results, as summarized in Table \ref{tab:3} and Table \ref{tab:4}, reveal several key insights into the efficacy of the proposed learning strategies and the nature of 4D affective perception.

While the Video-CNN and 4D-Vertex models initially struggled with raw data, the implementation of the difference-aware learning strategy led to a significant performance breakthrough. By supervising the models with residuals relative to the neutral state, both architectures achieved high correlation scores (PCC/CCC $>$ 0.9). This confirms that explicitly filtering out "static morphological noise" (identity-related geometry) is crucial. It allows the networks to transform from simple spatial feature extractors into sensitive detectors of dynamic deformation, mirroring the human visual system's ability to prioritize motion over static structure in expression perception.

% Notably, the 4D-Vertex model achieves an Intraclass Correlation Coefficient (ICC) of 0.975, which closely aligns with the reliability levels of human expert observers. This high alignment suggests that the subtle intensity variations encoded in 4D mesh sequences are not only objectively measurable but are also perceptually consistent with how humans interpret rendered video streams. The superior performance of 4D-Vertex over the FLAME-based baseline ($0.975$ vs. $0.946$ in ICC) indicates that raw vertex deformations capture more nuanced expressive details than a constrained parametric model.

Furthermore, a distinct performance gap is evident across modalities. Although Video-CNN captures temporal dynamics, it records the highest MAE ($0.289$), highlighting the inherent limitation of 2D projections in capturing fine-grained depth deformations. Conversely, the 4D-Vertex model achieves an exceptional ICC of $0.975$, closely aligning with the reliability levels of human expert observers. Its superior performance over both Video-CNN and the FLAME-based baseline ($0.139$ vs. $0.192$ in MAE) indicates that raw 4D vertex deformations encapsulate more comprehensive and nuanced expressive details than either 2D appearances or constrained parametric representations.

The results in Table \ref{tab:4} demonstrate the robust generalization capability of our approach. Despite being evaluated on a cross-identity set featuring unseen facial shapes and novel Action Unit (AU) combinations, the performance degradation is negligible (e.g., 4D-Vertex MAE only increases from 0.139 to 0.146). This stability indicates that the model has not merely memorized the probability distribution of the training data. Instead, it has learned a generalized motion paradigm—a fundamental mapping between geometric displacement and affective intensity that transcends individual identity.
\section{CONCLUSION}
\label{sec:conclusion}
In this paper, we present the 4D Facial Expression Intensity Dataset (4D-FEID), designed to support research on the perception of facial expression intensity. The dataset comprises 2,869 4D facial sequences, along with corresponding rendered videos at 60 FPS and more than 90,000 human-annotated intensity ratings.

% Furthermore, we benchmark the dataset using several representative deep networks and 4D aggregation methods, establishing baseline performance for facial expression intensity estimation. This resource provides a critical foundation for advancing affective computing, human behavior analysis, and the development of more natural and responsive human-computer interaction systems.
Furthermore, we benchmark the dataset using representative deep networks across different modalities. Our extensive evaluations demonstrate that leveraging dynamic 3D representations—particularly when coupled with difference-aware learning to filter out static morphological noise—significantly outperforms traditional 2D and parametric baselines. This confirms the necessity of high-fidelity 4D spatio-temporal modeling in capturing nuanced affective dynamics that align with human perception.

% While 4D-FEID establishes a robust foundation, future work will aim to address the reliance on parametric generation by incorporating in-the-wild 4D captures. Additionally, we plan to extend our analysis to explore cross-cultural variations in the perception of expression intensity and integrate psychophysical measurements to further investigate the cognitive thresholds of subtle facial deformations. Ultimately, this resource provides a critical stepping stone for advancing affective computing, human behavior analysis, and the development of more natural, responsive human-computer interaction systems.

\bibliographystyle{IEEEbib}
\bibliography{strings,refs}

@inproceedings{zhao2016facial,
  title={Facial expression intensity estimation using ordinal information},
  author={Zhao, Rui and Gan, Quan and Wang, Shangfei and Ji, Qiang},
  booktitle=CVPR,
  pages={3466--3474},
  year={2016}
}

@inproceedings{valstar2015fera,
  title={Fera 2015-second facial expression recognition and analysis challenge},
  author={Valstar, Michel F and Almaev, Timur and Girard, Jeffrey M and McKeown, Gary and Mehu, Marc and Yin, Lijun and Pantic, Maja and Cohn, Jeffrey F},
  booktitle=FG,
  volume={6},
  pages={1--8},
  year={2015}
}

@article{mavadati2013disfa,
  title={Disfa: A spontaneous facial action intensity database},
  author={Mavadati, S Mohammad and Mahoor, Mohammad H and Bartlett, Kevin and Trinh, Philip and Cohn, Jeffrey F},
  journal=TAFFC,
  volume={4},
  number={2},
  pages={151--160},
  year={2013},
  publisher={IEEE}
}

@article{zhang2014bp4d,
  title={Bp4d-spontaneous: a high-resolution spontaneous 3d dynamic facial expression database},
  author={Zhang, Xing and Yin, Lijun and Cohn, Jeffrey F and Canavan, Shaun and Reale, Michael and Horowitz, Andy and Liu, Peng and Girard, Jeffrey M},
  journal=IVC,
  volume={32},
  number={10},
  pages={692--706},
  year={2014},
  publisher=Elsevier
}

@inproceedings{zhang2013high,
  title={A high-resolution spontaneous 3d dynamic facial expression database},
  author={Zhang, Xing and Yin, Lijun and Cohn, Jeffrey F and Canavan, Shaun and Reale, Michael and Horowitz, Andy and Liu, Peng},
  booktitle=FG,
  pages={1--6},
  year={2013}
}

@article{ekman2002facial,
  title={The facial action coding system Second edition},
  author={Ekman, Paul and Friesen, Wallace V},
  journal=RNeBook,
  year={2002}
}

@inproceedings{COMAECCV18,
    title = {Generating {3D} faces using Convolutional Mesh Autoencoders},
    author = {Ranjan, Anurag and Bolkart, Timo and Sanyal, Soubhik and Black, Michael J},
    booktitle = ECCV,
    pages = {725--741},
    year = {2018}
}

@article{li2017learning,
  title={Learning a model of facial shape and expression from 4D scans},
  author={Li, Tianye and Bolkart, Timo and Black, Michael J and Li, Hao and Romero, Javier},
  journal=TOG,
  volume={36},
  number={6},
  pages={194--1},
  year={2017}
}

@article{muukkonen2020spatio,
  title={Spatio-temporal dynamics of face perception},
  author={Muukkonen, Ilkka and {\"O}lander, Kaisu and Numminen, Jussi and Salmela, Viljami R},
  journal={NeuroImage},
  volume={209},
  pages={116531},
  year={2020},
  publisher={Elsevier}
}

@article{ekman1987universals,
  title={Universals and cultural differences in the judgments of facial expressions of emotion},
  author={Ekman, Paul and Friesen, Wallace V and O'sullivan, Maureen and Chan, Anthony and Diacoyanni-Tarlatzis, Irene and Heider, Karl and Krause, Rainer and LeCompte, William Ayhan and Pitcairn, Tom and Ricci-Bitti, Pio E},
  journal=JPSP,
  volume={53},
  number={4},
  pages={712},
  year={1987},
  publisher=APA
}

@inproceedings{delannoy2008automatic,
  title={Automatic estimation of the dynamics of facial expression using a three-level model of intensity},
  author={Delannoy, Jane Reilly and McDonald, John},
  booktitle=FG,
  pages={1--6},
  year={2008}
}

@inproceedings{chang2013intensity,
  title={Intensity rank estimation of facial expressions based on a single image},
  author={Chang, Kuang-Yu and Chen, Chu-Song and Hung, Yi-Ping},
  booktitle=SMC,
  pages={3157--3162},
  year={2013}
}

@inproceedings{gudi2015deep,
  title={Deep learning based facs action unit occurrence and intensity estimation},
  author={Gudi, Amogh and Tasli, H Emrah and Den Uyl, Tim M and Maroulis, Andreas},
  booktitle=FG,
  volume={6},
  pages={1--5},
  year={2015}
}

@inproceedings{walecki2017deep,
  title={Deep structured learning for facial action unit intensity estimation},
  author={Walecki, Robert and Pavlovic, Vladimir and Schuller, Bj{\"o}rn and Pantic, Maja},
  booktitle=CVPR,
  pages={3405--3414},
  year={2017}
}

@inproceedings{fan2020facial,
  title={Facial action unit intensity estimation via semantic correspondence learning with dynamic graph convolution},
  author={Fan, Yingruo and Lam, Jacqueline and Li, Victor},
  booktitle=AAAI,
  volume={34},
  number={07},
  pages={12701--12708},
  year={2020}
}

@article{ma2024facial,
  title={Facial action unit detection and intensity estimation from self-supervised representation},
  author={Ma, Bowen and An, Rudong and Zhang, Wei and Ding, Yu and Zhao, Zeng and Zhang, Rongsheng and Lv, Tangjie and Fan, Changjie and Hu, Zhipeng},
  journal=TAFFC,
  volume={15},
  number={3},
  pages={1669--1683},
  year={2024},
  publisher={IEEE}
}

@inproceedings{mou2015group,
  title={Group-level arousal and valence recognition in static images: Face, body and context},
  author={Mou, Wenxuan and Celiktutan, Oya and Gunes, Hatice},
  booktitle=FG,
  volume={5},
  pages={1--6},
  year={2015}
}

@article{dhall2015automatic,
  title={Automatic group happiness intensity analysis},
  author={Dhall, Abhinav and Goecke, Roland and Gedeon, Tom},
  journal=TAFFC,
  volume={6},
  number={1},
  pages={13--26},
  year={2015},
  publisher={IEEE}
}

@article{li20224dme,
  title={4DME: A spontaneous 4D micro-expression dataset with multimodalities},
  author={Li, Xiaobai and Cheng, Shiyang and Li, Yante and Behzad, Muzammil and Shen, Jie and Zafeiriou, Stefanos and Pantic, Maja and Zhao, Guoying},
  journal=TAFFC,
  volume={14},
  number={4},
  pages={3031--3047},
  year={2022},
  publisher={IEEE}
}

@inproceedings{cosker2011facs,
  title={A FACS valid 3D dynamic action unit database with applications to 3D dynamic morphable facial modeling},
  author={Cosker, Darren and Krumhuber, Eva and Hilton, Adrian},
  booktitle=ICCV,
  pages={2296--2303},
  year={2011}
}

@inproceedings{yan2018spatial,
  title={Spatial temporal graph convolutional networks for skeleton-based action recognition},
  author={Yan, Sijie and Xiong, Yuanjun and Lin, Dahua},
  booktitle=AAAI,
  volume={32},
  number={1},
  year={2018}
}

@article{qi2017pointnet++,
  title={Pointnet++: Deep hierarchical feature learning on point sets in a metric space},
  author={Qi, Charles Ruizhongtai and Yi, Li and Su, Hao and Guibas, Leonidas J},
  journal=NeurIPS,
  volume={30},
  year={2017}
}

@article{li2022cas,
  title={CAS (ME) 3: A third generation facial spontaneous micro-expression database with depth information and high ecological validity},
  author={Li, Jingting and Dong, Zizhao and Lu, Shaoyuan and Wang, Su-Jing and Yan, Wen-Jing and Ma, Yinhuan and Liu, Ye and Huang, Changbing and Fu, Xiaolan},
  journal=TPAMI,
  volume={45},
  number={3},
  pages={2782--2800},
  year={2022},
  publisher={IEEE}
}

@book{cohen2013statistical,
  title={Statistical power analysis for the behavioral sciences},
  author={Cohen, Jacob},
  year={2013},
  publisher={routledge}
}

@misc{ITUP910,
  title        = {Subjective Video Quality Assessment Methods for Multimedia Applications},
  author       = {{ITU-T} {Rec.} {P.910}},
  year         = {2023},
  howpublished = {International Telecommunication Union, Technical Report}
}

@article{cheng20174dfab,
  title={4dfab: A large scale 4d facial expression database for biometric applications},
  author={Cheng, Shiyang and Kotsia, Irene and Pantic, Maja and Zafeiriou, Stefanos},
  journal={arXiv preprint arXiv:1712.01443},
  year={2017}
}

@inproceedings{principi2023florence,
  title={The florence 4d facial expression dataset},
  author={Principi, Filippo and Berretti, Stefano and Ferrari, Claudio and Otberdout, Naima and Daoudi, Mohamed and Del Bimbo, Alberto},
  booktitle=FG,
  pages={1--6},
  year={2023},
  organization={IEEE}
}

@string{Elsevier = "Elsevier"}

@string{APA = "American Psychological Association"}

@string{CVPR = "Proceedings of the IEEE Conference on Computer Vision and Pattern Recognition"}

@string{ICCV = "Proceedings of the IEEE International Conference on Computer Vision"}

@string{FG = "IEEE International Conference and Workshops on Automatic Face and Gesture Recognition"}

@string{SMC = "IEEE International Conference on Systems, Man, and Cybernetics"}

@string{AAAI = "Proceedings of the AAAI Conference on Artificial Intelligence"}

@string{TAFFC = "IEEE Transactions on Affective Computing"}

@string{TPAMI = "IEEE Transactions on Pattern Analysis and Machine Intelligence"}

@string{TOG = "ACM Transactions on Graphics"}

@string{IVC = "Image and Vision Computing"}

@string{JPSP = "Journal of Personality and Social Psychology"}

@string{RNeBook = "Salt Lake City: Research Nexus eBook"}

@string{CVPR = "Proc. IEEE Conf. Comput. Vis. Pattern Recognit."}

@string{ICCV = "Proc. IEEE Int. Conf. Comput. Vis."}

@string{FG = "IEEE Int. Conf. Automatic Face and Gesture Recognition"}

@string{SMC = "Proc. IEEE Int. Conf. Syst., Man, Cybern."}

@string{AAAI = "Proc. AAAI Conf. Artif. Intell."}

@string{ECCV = "European Conf. Comput. Vis."}

@string{TAFFC = "IEEE Trans. Affect. Comput."}

@string{TPAMI = "IEEE Trans. Pattern Anal. Mach. Intell."}

@string{NeurIPS = "Adv. Neural Inf. Process. Syst."}

@string{TOG = "ACM Trans. Graph."}

@string{IVC = "Image Vis. Comput."}

@string{JPSP = "J. Pers. Soc. Psychol."}

@string{ECCV = "European Conference on Computer Vision"}

@string{NeurIPS = "Advances in Neural Information Processing Systems"}
\end{document}